\documentclass[conference]{IEEEtran}
	
	\usepackage{amsmath,amssymb,amsfonts}
	\usepackage{graphicx}
	\graphicspath{{figures/}}
	\usepackage{cite}
	\usepackage{float}
	\usepackage{booktabs}
	\usepackage{multirow}
	\usepackage{url}
	\usepackage{mathtools}
	\usepackage{stfloats}
	\usepackage{cuted}
	\usepackage{threeparttable}
	\usepackage{xcolor}
	\usepackage{hyperref}
    \usepackage{booktabs}
    \usepackage{tabularx}
    \usepackage{makecell}

	\def\BibTeX{{\rm B\kern-.05em{\sc i\kern-.025em b}\kern-.08em
			T\kern-.1667em\lower.7ex\hbox{E}\kern-.125emX}}
		
	\hypersetup{
		colorlinks=true,
		citecolor=blue,
		linkcolor=blue,
		urlcolor=blue
	}
	
\begin{document}
		
	\title{STR Robot: Design of an Autonomous Mobile Robot from Simulation to Reality}
		
	\author{
		\IEEEauthorblockN{
			Vinh Nguyen\IEEEauthorrefmark{2},
			Gia-Uy Le\IEEEauthorrefmark{2},
			Tien-Dat Nguyen,
			Tri-Tin Nguyen,
			Vinh-Hao Nguyen\IEEEauthorrefmark{1}
		}
		\IEEEauthorblockA{
			Faculty of Electrical and Electronic Engineering, Ho Chi Minh City University of Technology, VNU-HCM \\ Ho Chi Minh City, Vietnam
		}
		\IEEEauthorblockA{
			vinh.nguyen2604@hcmut.edu.vn, uy.legia7@hcmut.edu.vn, nguyentiendat.sdh242@hcmut.edu.vn \\
			tin.nguyen2004@hcmut.edu.vn, vinhhao@hcmut.edu.vn
		}
		\IEEEauthorblockA{\IEEEauthorrefmark{2}These authors contributed equally to this work.}
		\IEEEauthorblockA{\IEEEauthorrefmark{1}Corresponding author.}
	}
		\maketitle
		
		\begin{abstract}
			With the rapid development of simulation tools, the development and validation of autonomous robotic systems have become more efficient before real-world deployment. This paper presents a simulation-to-real implementation of an autonomous mobile robot based on an existing mechanical platform. Instead of focusing on mechanical design, our work concentrates on the development of the onboard control, self-localization, and autonomous navigation system. The proposed robot is equipped with onboard sensing and computation to estimate its pose and navigate autonomously in the environment. The overall framework is first developed and tested in simulation, and then deployed on the real robot for experimental evaluation. The results demonstrate the feasibility of the proposed approach and show that simulation provides an effective foundation for developing reliable autonomous mobile robot systems. The source code will be released at \url{https://ntdathp.github.io/outdoor-robot-web}.
		\end{abstract}
			
		
	\section{Introduction}

    Autonomous mobile robots are increasingly used in environmental monitoring, inspection, logistics, and service robotics. In outdoor environments, reliable navigation requires the real-time integration of localization, path planning, and trajectory tracking under sensing uncertainty, model mismatch, and limited computational resources. Therefore, practical autonomy should be developed as a unified onboard system rather than as separate modules~\cite{kim2022opensource_nav,yu2018bus_nav}.
    
    Testing autonomous navigation directly on physical robots is costly, time-consuming, and potentially unsafe in early development. Simulation provides a safer and more repeatable way to prototype and evaluate the system before real-world deployment. However, transferring a navigation framework from simulation to an outdoor robot remains challenging due to sensing differences, environmental complexity, and imperfect modeling~\cite{wiedemann2024simmodel}.
    
    Motivated by these challenges, this paper presents a simulation-to-real autonomous navigation system for an existing outdoor mobile robot platform. The proposed framework focuses on the onboard autonomy stack, including self-localization, map-based path planning, and an Ackermann Geometric MPC (A-GMPC) tracking controller. The system is first developed and evaluated in simulation, then deployed on the real robot for outdoor experiments.
    
    The main objective is to establish a practical and reproducible simulation-to-real pipeline while examining the behavior of modern planning and tracking modules in a complete onboard navigation system. In particular, this work investigates the integration of a geometric tracking formulation for an Ackermann-steered platform within a real-time localization--planning--control pipeline.
    
    \begin{figure}[t]
        \centering
        \includegraphics[width=\linewidth]{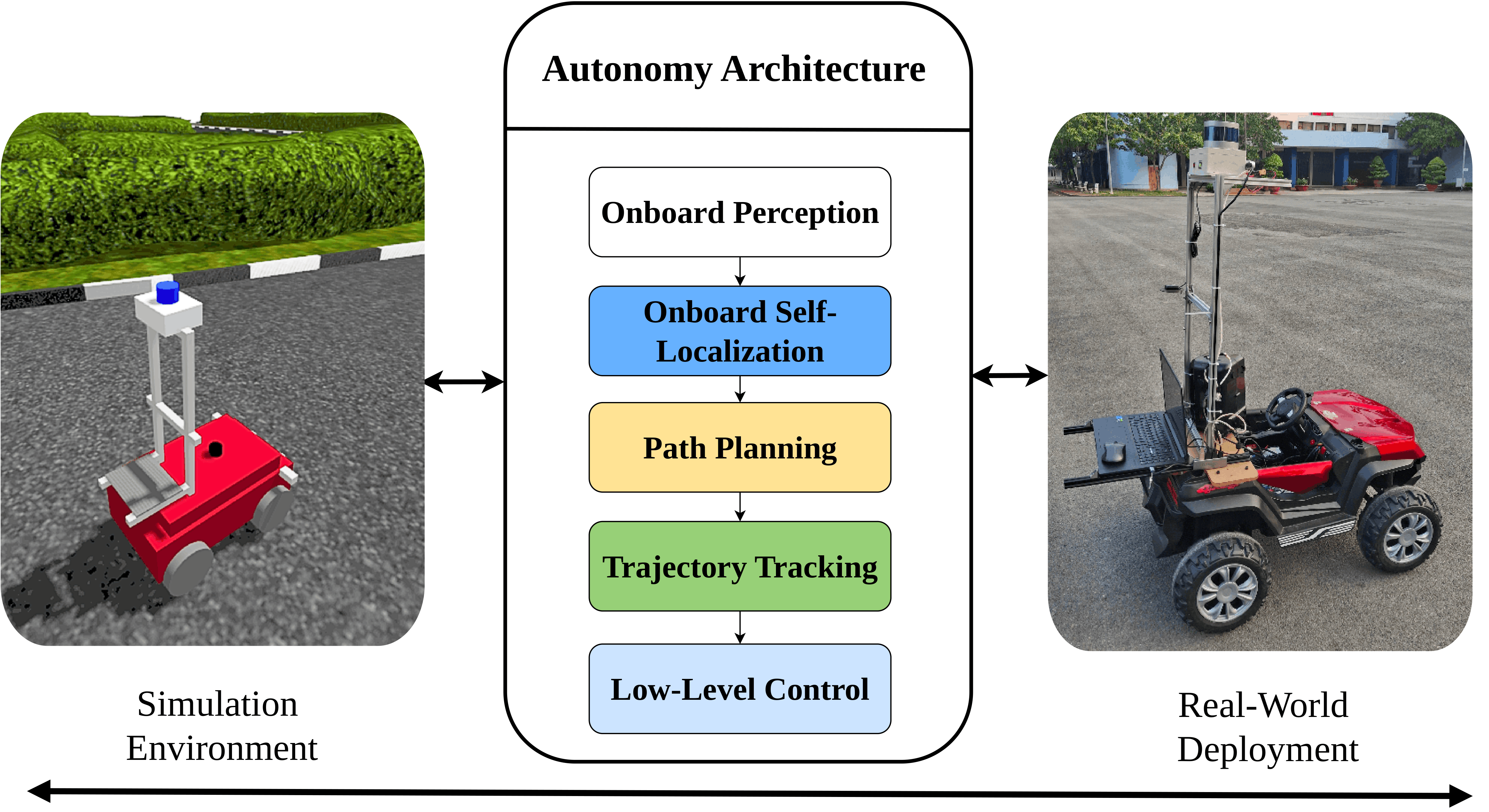}
        \caption{Simulation-to-real workflow of the proposed autonomous mobile robot, illustrating the autonomy stack bridging simulation and real-world deployment.}
        \label{fig:sim2real_framework}
    \end{figure}
    
    The main contributions of this paper are summarized as follows:
    \begin{itemize}
        \item A complete simulation-to-real navigation framework for an outdoor autonomous mobile robot, integrating onboard localization, path planning, and path tracking.
        \item The integration of an Ackermann Geometric MPC (A-GMPC) tracking controller into a unified onboard autonomy stack for both simulation and real-world experiments.
        \item The experimental setup, source code, and results are released for the benefit of the community.
    \end{itemize}

	\section{Related Work}
	
	\subsection{Path Planning}
	
Path planning is essential for autonomous navigation. Although classical $A^*$ ~\cite{hart1968astar} is efficient, it often produces paths close to obstacles with sharp turns, which is unsuitable for non-holonomic robots. Improved methods such as geometric $A^*$~\cite{tang2021geometric} enhance smoothness but are mainly designed for structured environments. Inspired by Lin \textit{et al.}~\cite{lin2024astar}, this work adopts a cost-aware planning strategy to generate smooth and collision-free reference paths for MPC tracking.

\subsection{Path Tracking}
\label{subsec:rel_path_tracking}
Model Predictive Control (MPC) is widely adopted for path tracking as it explicitly handles system constraints~\cite{le2024mpc}. Beyond linearized Euclidean models, nonlinear MPC approaches using vector-space or Euler-angle coordinates have also been widely investigated~\cite{jian2023mpc, song2023isolating}. However, these formulations typically rely on strict temporal references, which often induce oscillatory behavior or corner-cutting during sharp maneuvers.

To improve spatial consistency, recent approaches formulate tracking errors directly on configuration manifolds, such as the $SE(2)$ Geometric MPC (GMPC) framework~\cite{tang2024gmpc, lu2023ommpc}. Building on these developments, this work adopts the GMPC geometric error formulation and extends it into an Ackermann-oriented tracking controller. By integrating an input-increment optimization, the proposed method explicitly penalizes aggressive control actions to ensure smoother steering for practical real-world deployment.

\subsection{Onboard Localization and 3D Mapping}

Reliable onboard perception is crucial for autonomous outdoor navigation. While methods such as SLICT~\cite{nguyen2023slict} provide highly accurate continuous-time localization and mapping, their computational complexity makes real-time deployment on embedded robotic platforms difficult. In contrast, FAST-LIO2~\cite{xu2022fastlio2} offers an efficient LiDAR--inertial solution for real-time state estimation and 3D mapping. Building on this, FAST-LIVO2~\cite{zheng2025fastlivo2} further incorporates visual information to improve robustness and map quality when camera observations are informative. Therefore, in this work, FAST-LIVO2 is employed as the primary onboard localization and 3D mapping framework under normal illumination, while FAST-LIO2 is considered a practical alternative in dark or visually degraded environments where visual information becomes less reliable.

		\section{Methodology}

\subsection{System Overview}
\begin{figure}[htbp]
	\centering
	\includegraphics[width=0.95\linewidth]{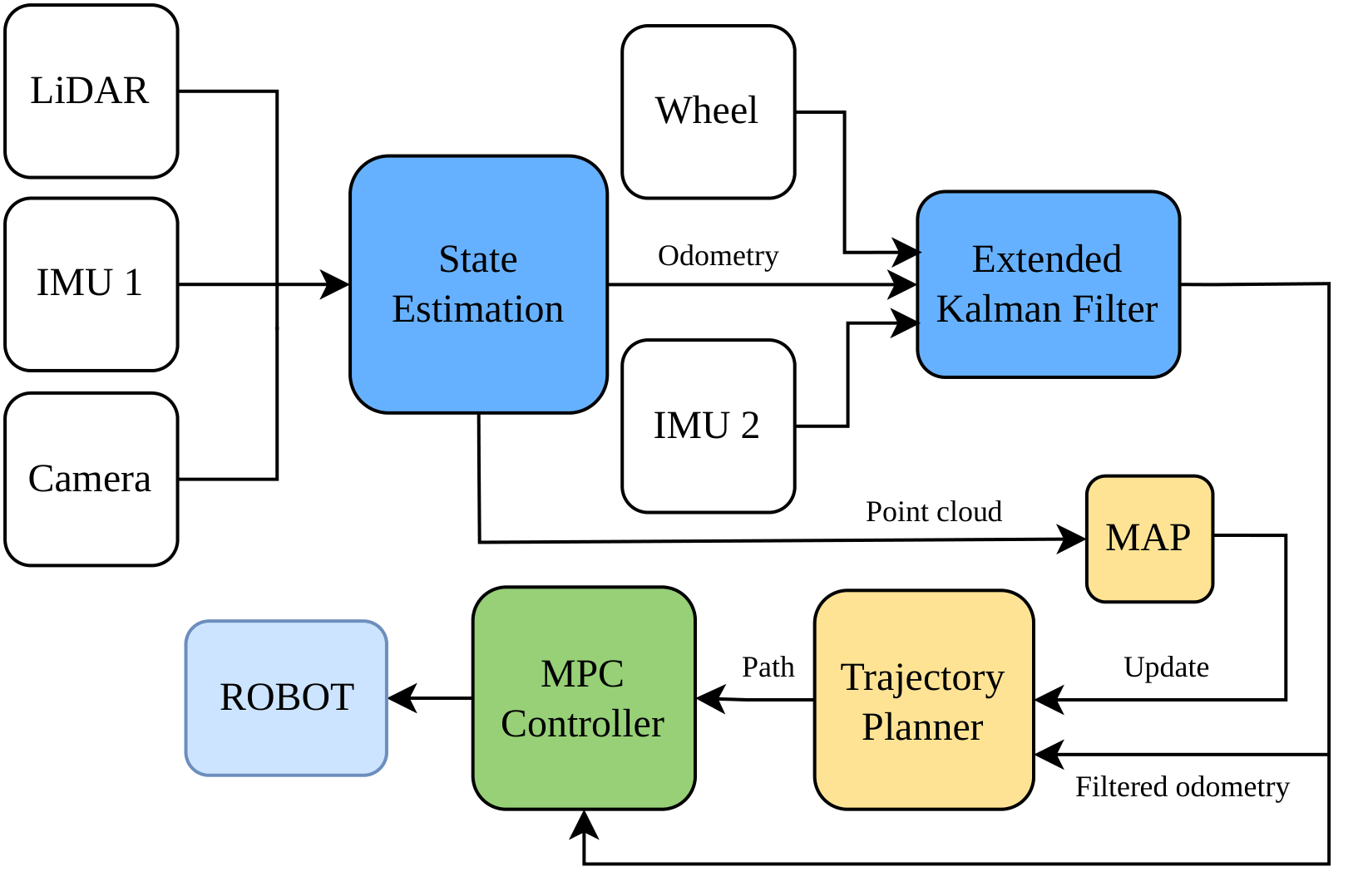}
	\caption{Overview of the proposed autonomous navigation framework. Multi-sensor odometry is fused by an EKF for state estimation, while the map and estimated state are used for trajectory planning and MPC-based path tracking.}
	\label{fig:system_overview}
\end{figure}

The proposed autonomous navigation framework consists of three main modules: onboard localization and mapping, global path planning, and path tracking, as illustrated in Fig.~\ref{fig:system_overview}. 
The onboard self-localization and mapping module performs real-time state estimation and reconstructs a registered 3D point-cloud map from LiDAR, IMU, and camera measurements. Its formulation is described in Section~\ref{subsec:localization}.
For ground navigation, the robot's pose is simplified to planar motion, and the 3D map is projected into a 2D occupancy map for navigation.
Using these inputs, the global planner generates a collision-free path from start to goal (Section~\ref{subsec:planning}), which can be further refined by the local obstacle avoidance module during execution for improved safety.

The resulting trajectory is tracked by the control module, where both the MPC baseline and the proposed A-GMPC are evaluated. 
Based on the reference trajectory and the estimated state, the controller computes online inputs for path tracking, as detailed in Section~\ref{subsec:tracking}.
\vspace{-2mm}
\subsection{Path Planning Module}
\label{subsec:planning}

The environment is represented as a 2D grid map, where each cell is classified as free or occupied. Given a start node $n_s$ and a goal node $n_g$, a collision-free path is generated.

To ensure safety clearance, obstacles are expanded as
\begin{equation}
	n_{ex} = \left\lceil \frac{D_s}{p} \right\rceil,
\end{equation}
where $D_s$ is the maximum distance from the vehicle center to its boundary and $p$ is the grid resolution.

Path planning is performed using an improved A* algorithm with an evaluation function
\begin{equation}
	F(n) = G(n) + H(n),
\end{equation}
where $G(n)$ denotes the accumulated cost from the start node to node $n$, and the heuristic is defined as the Euclidean distance:
\begin{equation}
	H(n) = \sqrt{(x_n - x_g)^2 + (y_n - y_g)^2}.
\end{equation}

The step cost between two adjacent nodes is defined as
\begin{equation}
	P(n) = \sqrt{(x_n - x_{n-1})^2 + (y_n - y_{n-1})^2}.
\end{equation}

The cost function is modified to penalize direction changes:
\begin{equation}
	G(n)=
	\begin{cases}
		G(n-1) + P(n), & \text{if collinear},\\
		G(n-1) + P(n) + C, & \text{otherwise},
	\end{cases}
\end{equation}
where $C$ is the turning cost.

To improve computational efficiency, the algorithm checks at each expansion step whether the current node can be directly connected to the goal without collision. If so, the search terminates early.

After obtaining the discrete grid-based path, redundant inflection points are removed. A node $n_k$ is eliminated if three consecutive nodes are collinear:
\begin{equation}
	\overline{n_{k-1}n_k} \parallel \overline{n_kn_{k+1}}.
\end{equation}

This post-processing step reduces unnecessary direction changes and improves path smoothness.

Finally, the global path is simplified and smoothed to generate a reference trajectory
\begin{equation}
	\{T_{ref,k}, u_{ref,k}\}_{k=0}^{N}, \quad T_{ref,k} \in SE(2),
\end{equation}
where $T_{ref,k}$ is the reference pose and $u_{ref,k}$ is the corresponding control input.

During execution, dynamic obstacle avoidance is performed locally by monitoring the global path within the local costmap. When a moving obstacle is detected along the current path segment, the affected portion is replaced by a $C^2$-smooth curve
\begin{equation}
	B(t) = (1-t)^3 P_0 + 3(1-t)^2 t P_1 + 3(1-t)t^2 P_2 + t^3 P_3,
\end{equation}
with $t \in [0,1]$. Here, $P_0$ and $P_3$ are fixed on the original path to ensure smooth re-entry, while $P_1$ and $P_2$ are shifted along the normal direction to maintain a safe clearance from the obstacle. This local update preserves the global route while enabling real-time avoidance of moving obstacles.
	\subsection{Path Tracking Module}
	\label{subsec:tracking}
	
\begin{figure}[htbp]
		\centering
		\includegraphics[width=0.85\linewidth, trim={0cm 2.0cm 0cm 2cm}, clip]{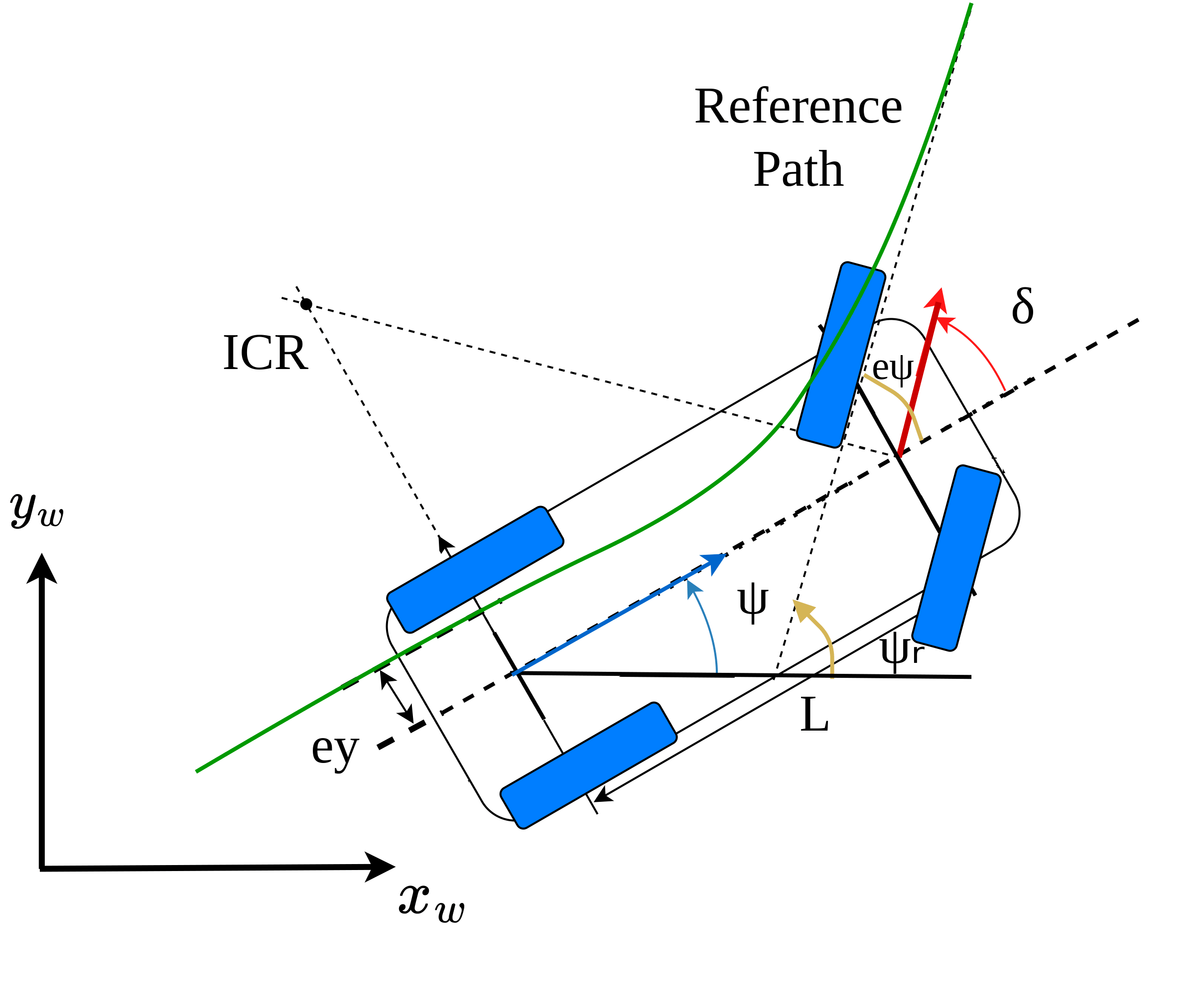}
		\caption{Ackermann kinematic bicycle model and tracking-error variables used in the baseline formulation~\cite{le2024mpc}.}
		\label{fig:kinematic_model}
	\end{figure}

	For trajectory tracking, two controllers are implemented for comparative evaluation: a standard MPC baseline and the proposed Ackermann Geometric MPC (A-GMPC). The standard MPC baseline adopts the linearized Euclidean error-state formulation and the Ackermann kinematic bicycle model (illustrated in Fig.~\ref{fig:kinematic_model}) presented in~\cite{le2024mpc}. It solves a finite-horizon quadratic optimization problem to minimize local lateral and heading errors subject to physical steering limits.
	
	Building upon the geometric tracking concepts in Section~\ref{subsec:rel_path_tracking}, the proposed A-GMPC defines the tracking error directly on the robot pose manifold $SE(2)$. Let $T, T_{\mathrm{ref}} \in SE(2)$ denote the current and reference poses. The geometric tracking error $E \in SE(2)$ and its Lie algebra representation $\xi \in \mathbb{R}^3$ are defined as:
	\begin{equation}
		E = T_{\mathrm{ref}}^{-1}T, \quad \xi = \mathrm{Log}(E) = \begin{bmatrix} e_x & e_y & e_\psi \end{bmatrix}^{T}.
	\end{equation}
	Defining the control input $u=[v, \omega]^{T}$, the reference input $u_{\mathrm{ref},k}$, and the input deviation $\tilde{u}_k = u_k - u_{\mathrm{ref},k}$, the continuous-time error dynamics linearized around the reference trajectory are given by~\cite{tang2024gmpc} as:
	\begin{equation}
		\dot{\xi} = A_{c,k}\xi + B_c\tilde{u}_k,
	\end{equation}
	where
	\begin{equation}
		A_{c,k} =
		\begin{bmatrix}
			0 & \omega_{\mathrm{ref},k} & 0\\
			-\omega_{\mathrm{ref},k} & 0 & v_{\mathrm{ref},k}\\
			0 & 0 & 0
		\end{bmatrix},
		\quad
		B_c =
		\begin{bmatrix}
			1 & 0\\
			0 & 0\\
			0 & 1
		\end{bmatrix}.
	\end{equation}
	
	Using Euler discretization with sampling time $\Delta t$, the discrete-time dynamics become:
	\begin{equation}
		\xi_{k+1} = A_{d,k}\xi_k + B_{d,k}\tilde{u}_k,
	\end{equation}
	where $A_{d,k} = I + A_{c,k}\Delta t$ and $B_{d,k} = B_c\Delta t$.
	
	While standard formulations optimize the control input directly, highly varying commands may lead to less smooth actuation on physical Ackermann steering systems. To penalize aggressive steering variations and account for actuator constraints, A-GMPC optimizes the input increment:
	\begin{equation}
		\Delta u_k = \tilde{u}_k - \tilde{u}_{k-1}.
	\end{equation}
	By introducing the augmented state $\bar{\xi}_k = [\xi_k^T, \tilde{u}_{k-1}^T]^T$, the augmented prediction model is written as:
	\begin{equation}
		\bar{\xi}_{k+1} =
		\begin{bmatrix}
			A_{d,k} & B_{d,k}\\
			0 & I
		\end{bmatrix} \bar{\xi}_k
		+
		\begin{bmatrix}
			B_{d,k}\\
			I
		\end{bmatrix} \Delta u_k.
	\end{equation}
	
	Based on this augmented prediction model, the controller solves the following finite-horizon optimization problem:
	\begin{equation}
		\min_{\{\Delta u_k\}}
		\sum_{k=0}^{N_p}
		\left(
		\|\xi_k\|_{Q}^{2} + \|\tilde{u}_k\|_{R}^{2}
		\right)
		+
		\sum_{k=0}^{N_p-1}
		\|\Delta u_k\|_{R_{\Delta}}^{2},
	\end{equation}
	subject to the augmented dynamics and the actuator bounds ($0 \le v_k \le v_{\max}$ and $|\omega_k| \le \omega_{\max}$). The optimized $(v,\omega)$ commands are subsequently converted to the Ackermann steering interface using:
	\begin{equation}
		\delta = \arctan\left(\frac{\omega L}{v}\right).
	\end{equation}
	By combining the geometric error formulation on $SE(2)$~\cite{tang2024gmpc} with augmented input-increment optimization, A-GMPC improves spatial tracking consistency while promoting smoother actuation for real-world deployment.
    
\subsection{Onboard Self-Localization and Mapping Module}
\label{subsec:localization}

The onboard self-localization (OSL) module adopts a dual-layer estimation architecture to provide high-rate feedback for control and globally consistent poses for navigation. Let $B$, $L$, and $M$ denote the robot body frame, local odom frame, and global map frame, respectively. At time index $k$, ${}^{L}_{B}\mathbf{x}_{k}$ and ${}^{M}_{B}\mathbf{x}_{k}$ denote the robot state in the local and map frames.

Scan-based odometry, such as LiDAR--inertial or LiDAR--inertial--visual odometry, is typically available at the LiDAR scan rate, approximately $10~\mathrm{Hz}$. However, the tracking controller requires a higher feedback rate, usually above $30~\mathrm{Hz}$. Therefore, a local EKF fuses low-rate scan-based odometry with high-rate wheel encoder and IMU measurements~\cite{moore2016ekf}:
\begin{equation}
    {}^{L}_{B}\mathbf{x}_{k}
    =
    f_{L}\left({}^{L}_{B}\mathbf{x}_{k-1}, \mathbf{u}_{k}\right)
    + {}^{L}\mathbf{w}_{k},
\end{equation}
\begin{equation}
    {}^{L}\mathbf{z}_{k}
    =
    h_{L}\left({}^{L}_{B}\mathbf{x}_{k}\right)
    + {}^{L}\mathbf{v}_{k},
\end{equation}
where $\mathbf{u}_{k}$ is the control input, ${}^{L}\mathbf{z}_{k}$ includes scan-based odometry, wheel odometry, and IMU measurements, and ${}^{L}\mathbf{w}_{k}$ and ${}^{L}\mathbf{v}_{k}$ are the process and measurement noise. Since the encoder and IMU operate at higher rates than LiDAR odometry, this layer provides smooth high-frequency state feedback for the controller.

At the map level, a second estimator maintains global consistency by correcting the local odom frame with respect to the map frame. ICP-based scan-to-map matching against a 2D occupancy map provides a map-referenced observation:
\begin{equation}
    {}^{M}\mathbf{z}_{k}
    =
    h_{\mathrm{ICP}}\left({}^{L}_{B}\mathbf{x}_{k}\right)
    + {}^{M}\mathbf{v}_{k},
\end{equation}
where ${}^{M}\mathbf{z}_{k}$ is the scan-to-map pose correction and ${}^{M}\mathbf{v}_{k}$ is the corresponding noise. The corrected robot pose is obtained as
\begin{equation}
    {}^{M}_{B}\mathbf{T}_{k}
    =
    {}^{M}_{L}\mathbf{T}_{k}
    {}^{L}_{B}\mathbf{T}_{k},
\end{equation}
where ${}^{M}_{L}\mathbf{T}_{k}$ is the transformation from the local odom frame to the global map frame. For planar navigation, the pose is projected to $(x,y,\theta)$ and used by the planner and tracking controller.

In addition to localization, the system builds a registered 3D point-cloud map ${}^{M}\mathcal{P}$. For ground navigation, the map is height-filtered as
\begin{equation}
    {}^{M}\mathcal{P}_{f}
    =
    \left\{
    {}^{M}\mathbf{p}_{i}\in{}^{M}\mathcal{P}
    \mid
    z_{\min}\le z_{i}\le z_{\max}
    \right\},
\end{equation}
and then projected onto a 2D occupancy grid. The module supports both offline mapping with a pre-built map and online mapping for previously unknown environments.

\section{Experiments}
		
The experiments were conducted in both simulation and real-world settings to evaluate the proposed system in path planning, path tracking, and autonomous navigation.
\vspace{-2mm}
	\subsection{Simulation Experiments}
	
	\subsubsection{Simulation Setup}
	
\begin{figure}[htbp]
		\centering
			\includegraphics[width=0.9\linewidth]{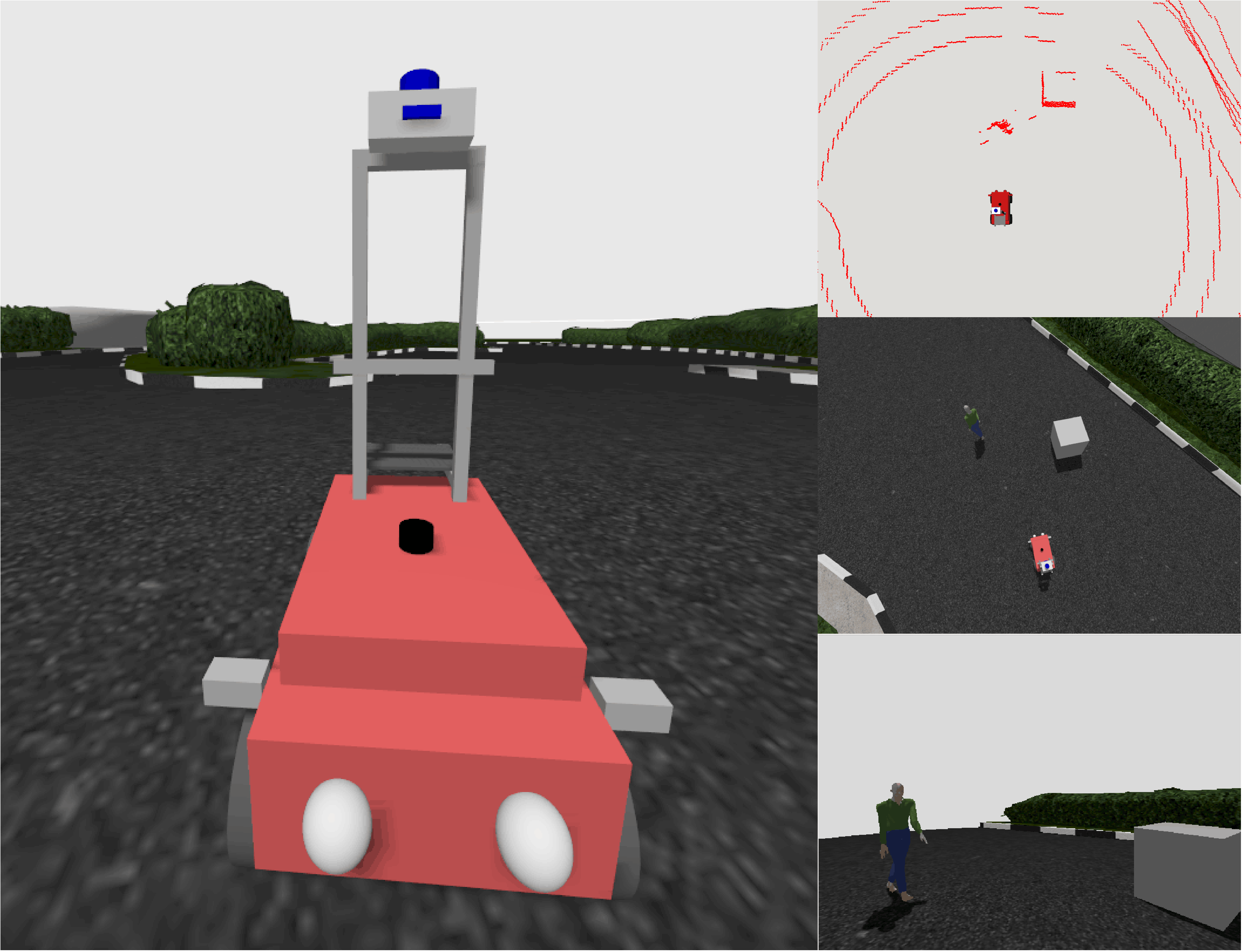}
			\caption{Simulation environment used for sim-to-real validation. The large panel shows the robot platform in the main scene, while the three smaller panels illustrate the LiDAR view, the top-down view, and the camera view of the simulated environment.}
		\label{fig:sim_env}
	\end{figure}

	The navigation architecture is validated in Gazebo using a non-holonomic robot model. For sim-to-real validation, the virtual environment shown in Fig.~\ref{fig:sim_env} is constructed to closely resemble the real-world experimental setting.
	\vspace{-0.2cm}

	\subsubsection{Path Planning Evaluation}
	
\begin{figure}[htbp]
		\centering
		\includegraphics[width=0.85\linewidth]{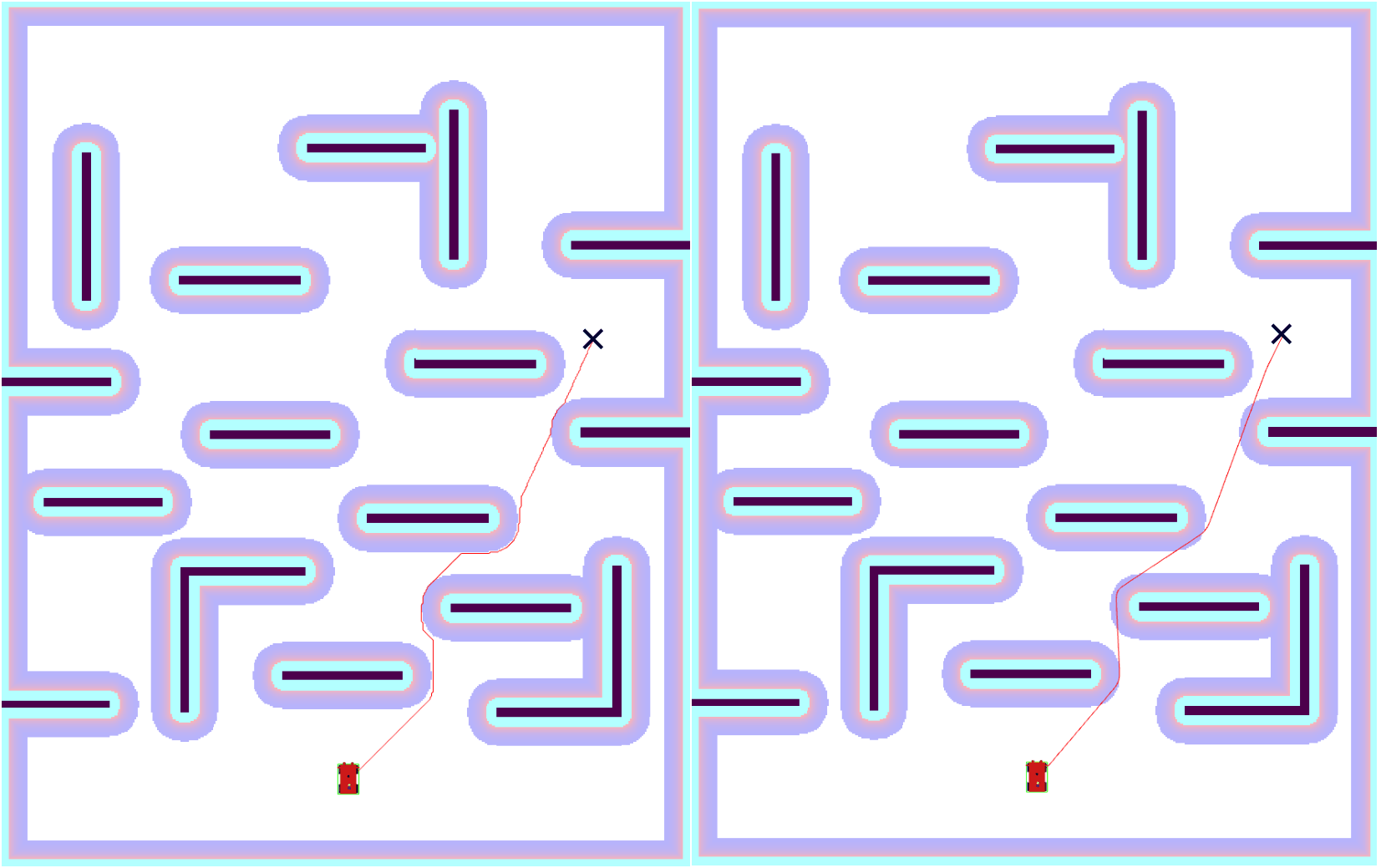}
		\caption{Qualitative comparison of the planned paths generated by the two algorithms in the same environment. The baseline A* planner (left) produces a path with multiple unnecessary turns and segments close to obstacle boundaries, whereas the improved A* planner (right) generates a smoother path with better clearance and fewer abrupt direction changes.}
		\label{fig:path_comparison}
	\end{figure}

	Based on the method in Section~\ref{subsec:planning}, the planner was evaluated in simulation on 20 randomized start--goal queries. 
	Table~\ref{tab:sim_planning_comparison} and Fig.~\ref{fig:path_comparison} show that the improved $A^*$ produces shorter and smoother paths with fewer direction changes, at the cost of slightly higher planning time. 
	A supplementary video of the path generation process is available at \url{https://youtu.be/LbHl0L6YVQk}.
	\vspace{-0.3cm}
	\begin{figure*}[h]
		\centering
		\begin{minipage}{0.24\textwidth}
			\centering
			\includegraphics[width=\linewidth]{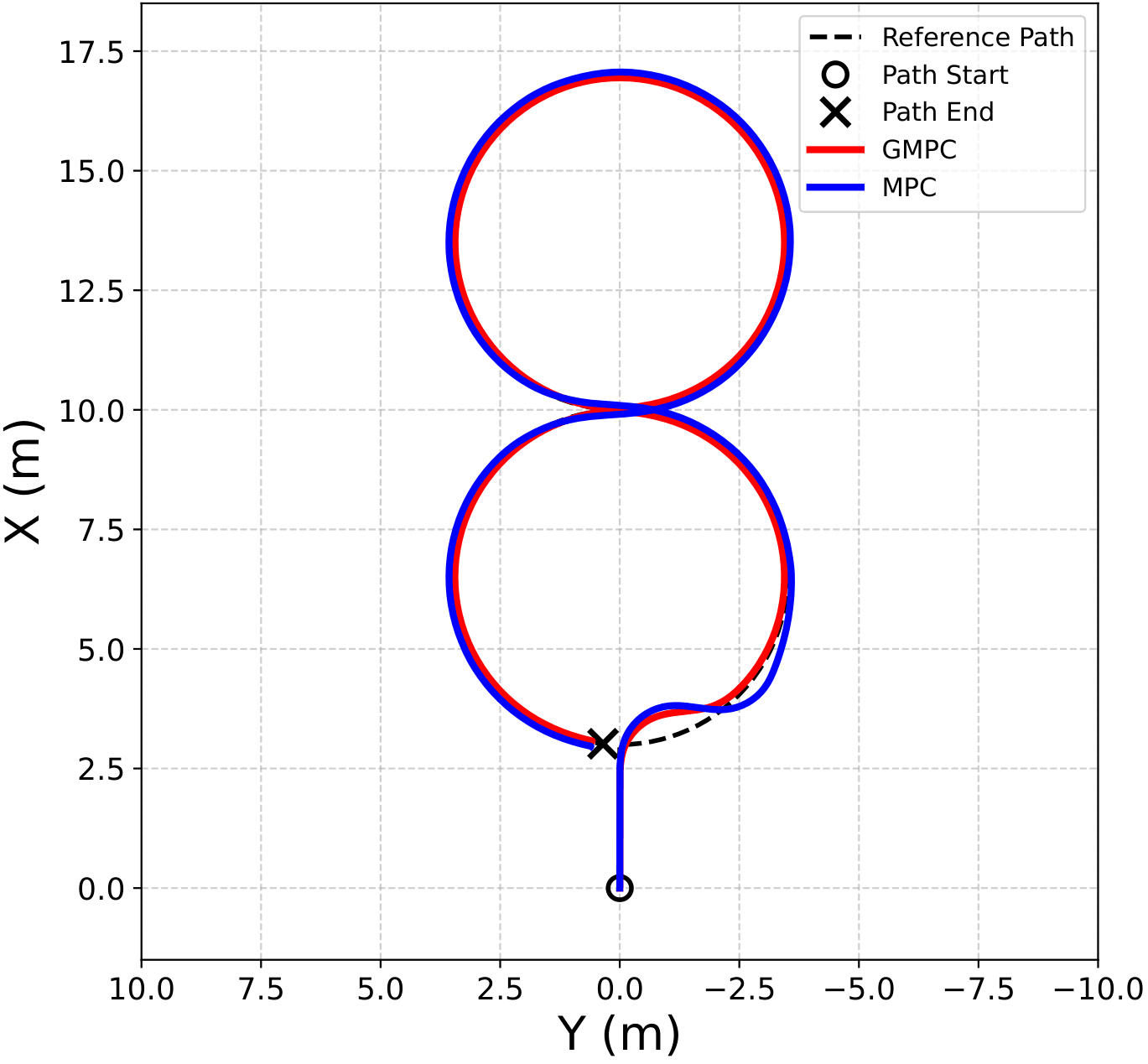}
			\centerline{(a) Sim: Lemniscate}
		\end{minipage}
		\hfill
		\begin{minipage}{0.24\textwidth}
			\centering
			\includegraphics[width=\linewidth]{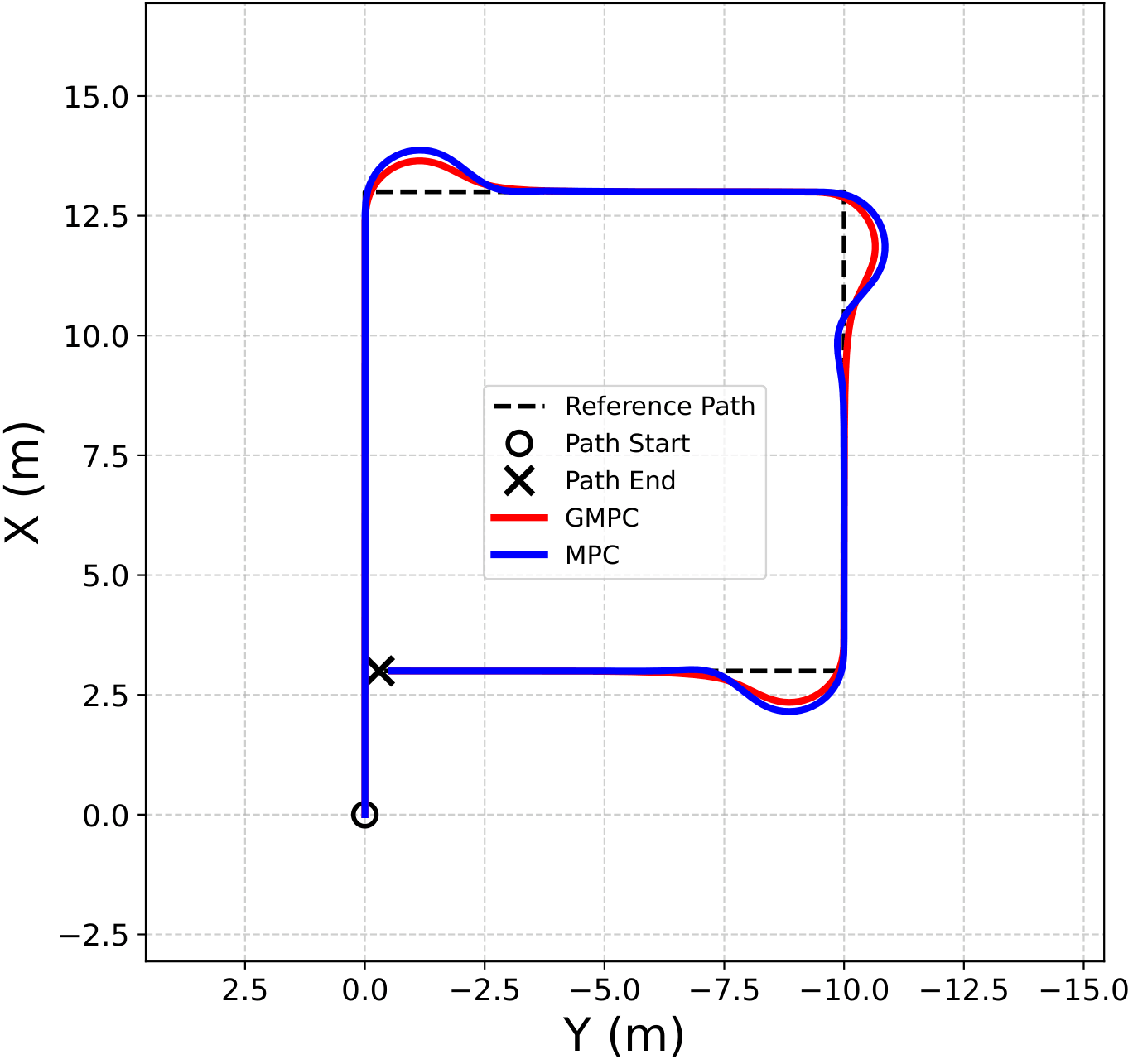}
			\centerline{(b) Sim: Square}
		\end{minipage}
		\hfill
		\begin{minipage}{0.24\textwidth}
			\centering
			\includegraphics[width=\linewidth]{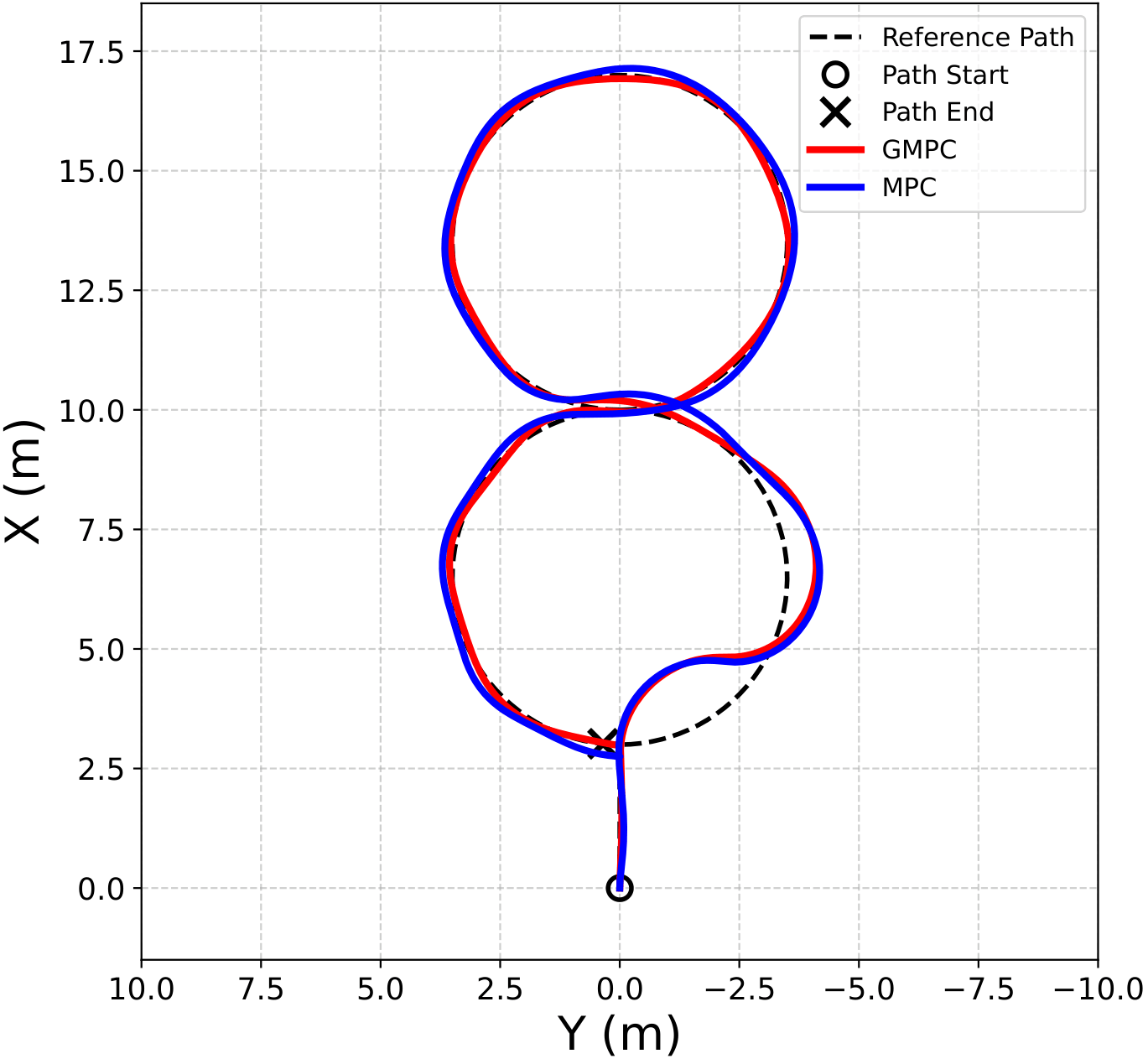}
			\centerline{(c) Real: Lemniscate}
		\end{minipage}
		\hfill
		\begin{minipage}{0.24\textwidth}
			\centering
			\includegraphics[width=\linewidth]{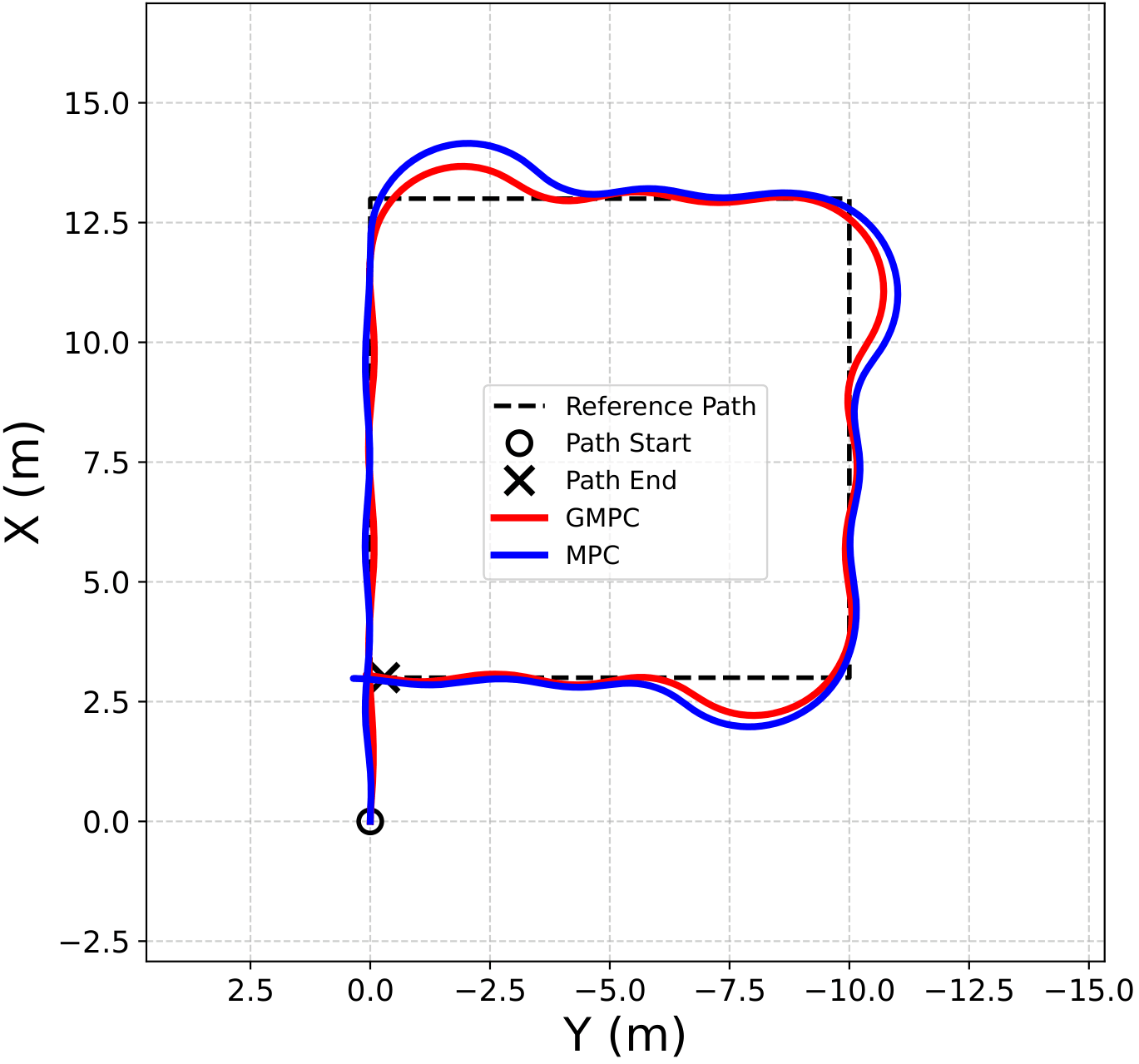}
			\centerline{(d) Real: Square}
		\end{minipage}
		\caption{Comparison of path-tracking performance between the standard MPC (blue) and the proposed A-GMPC (red). (a)--(b) Simulation results. (c)--(d) Representative real-world results at an average speed of approximately 1.1~m/s. Both controllers are able to track the continuous lemniscate path. On the square trajectory, the different cornering behaviors highlight the effect of their respective error-state formulations, with the proposed A-GMPC producing more accurate tracking in the real-world experiments.}
		\label{fig:all_tracking_paths}
	\end{figure*}
	
	\begin{table}[htbp]
		\centering
		\caption{Path planning performance comparison between baseline and improved $A^*$. The Best Results are in Bold. }
		\label{tab:sim_planning_comparison}
		\begin{tabular}{lcc}
			\toprule
			\textbf{Metric} & \textbf{Baseline $A^*$} & \textbf{Improved $A^*$} \\
			\midrule
			Path length (m)                & 25.52  & \textbf{22.35} \\
			Total turning angle ($^\circ$) & 3712.50 & \textbf{170.55} \\
			Number of inflection points    & 79.45  & \textbf{4.55} \\
			Planning time (ms)             & \textbf{221.15} & 330.54 \\
			\bottomrule
		\end{tabular}
	\end{table}
	
	\subsubsection{Path-Tracking Validation in Simulation}
	The tracking performance was evaluated on lemniscate and square trajectories. Table~\ref{tab:sim_tracking_rmse} and Fig.~\ref{fig:all_tracking_paths}(a)--(b) present the quantitative and qualitative results, respectively. A-GMPC achieves lower RMSE than the standard MPC on both trajectories, with a more noticeable improvement on the lemniscate path.
	
	\begin{table}[!htbp]
		\centering
		\caption{RMSE Comparison in Simulation (Unit [m]). The Best Results are in Bold.}
		\label{tab:sim_tracking_rmse}
		\begin{tabular}{lcc}
			\toprule 
			\textbf{Trajectory} & \textbf{Standard MPC} & \textbf{Proposed A-GMPC} \\
			\midrule
			Lemniscate & 0.1215 & \textbf{0.0764} \\
			Square     & 0.2427 & \textbf{0.1944} \\
			\bottomrule
		\end{tabular}
	\end{table}

	\subsubsection{Dynamic Obstacle Avoidance}
	
	Dynamic obstacle avoidance was evaluated under a hierarchical navigation framework, where the global $A^*$ path was generated once, and a local avoidance module updated the reference path online to react to moving obstacles. This allowed the robot to preserve the global route while adapting locally for collision avoidance. A supplementary video is available at \url{https://youtu.be/u9p_GGSCZNM}.
    
	\subsection{Real-World Experiments}
    
\begin{figure}
		\centering
        \vspace{-10mm}
		\includegraphics[width=\linewidth]{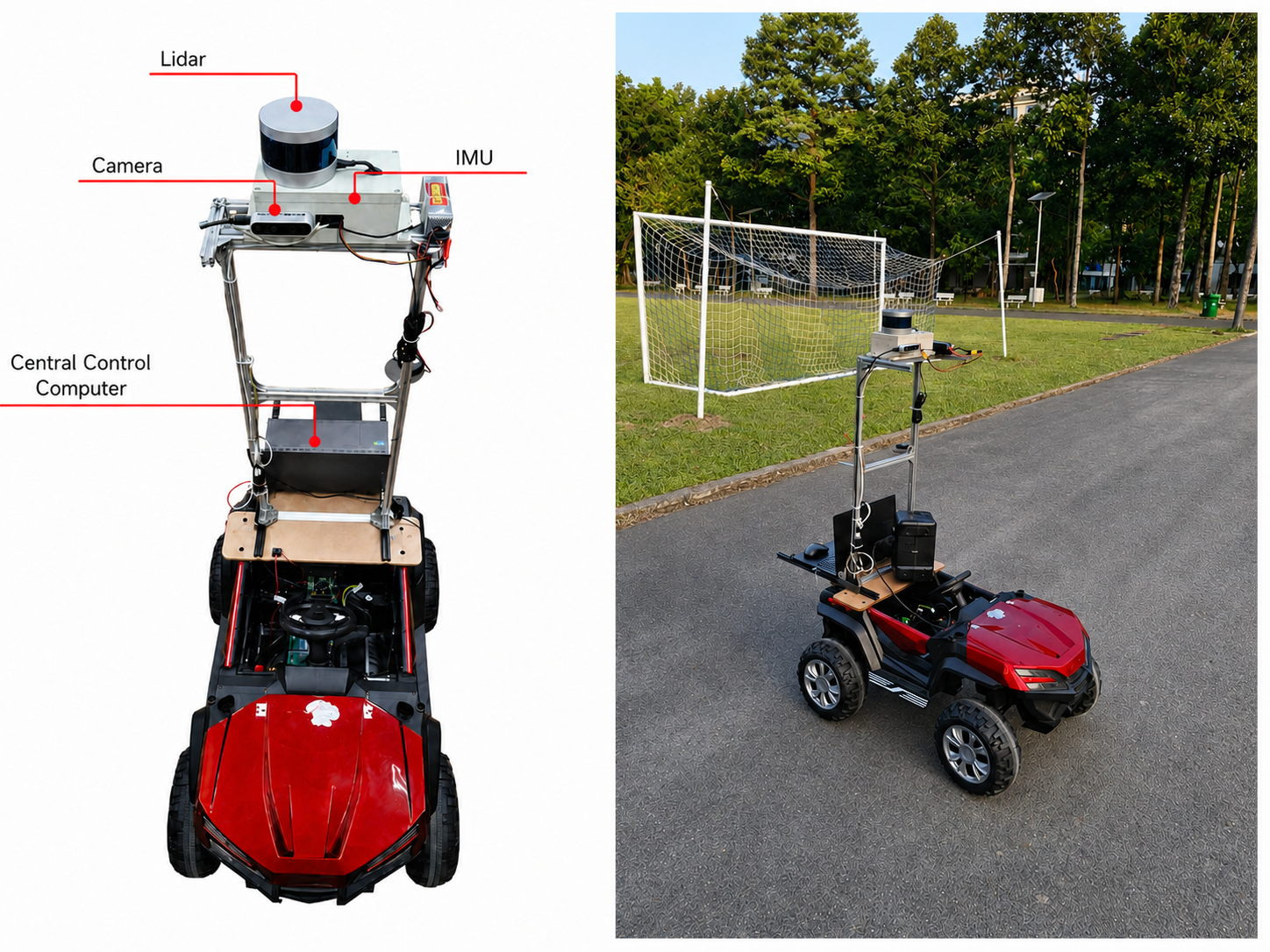}
        \vspace{-8mm}
		\caption{The physical Ackermann-steered mobile robot utilized for real-world validation. The sensor suite comprises a top-mounted 3D LiDAR, a forward-facing RGB-D camera, and an internal IMU, all processed by a central onboard computer for real-time autonomy.}
		\label{fig:real_robot}
        \vspace{-2mm}
	\end{figure}

	To validate the proposed framework in real-world conditions, experiments were conducted using the custom Ackermann-steered mobile robot shown in Fig.~\ref{fig:real_robot}.

	\begin{figure}
		\centering
		\includegraphics[width=\linewidth]{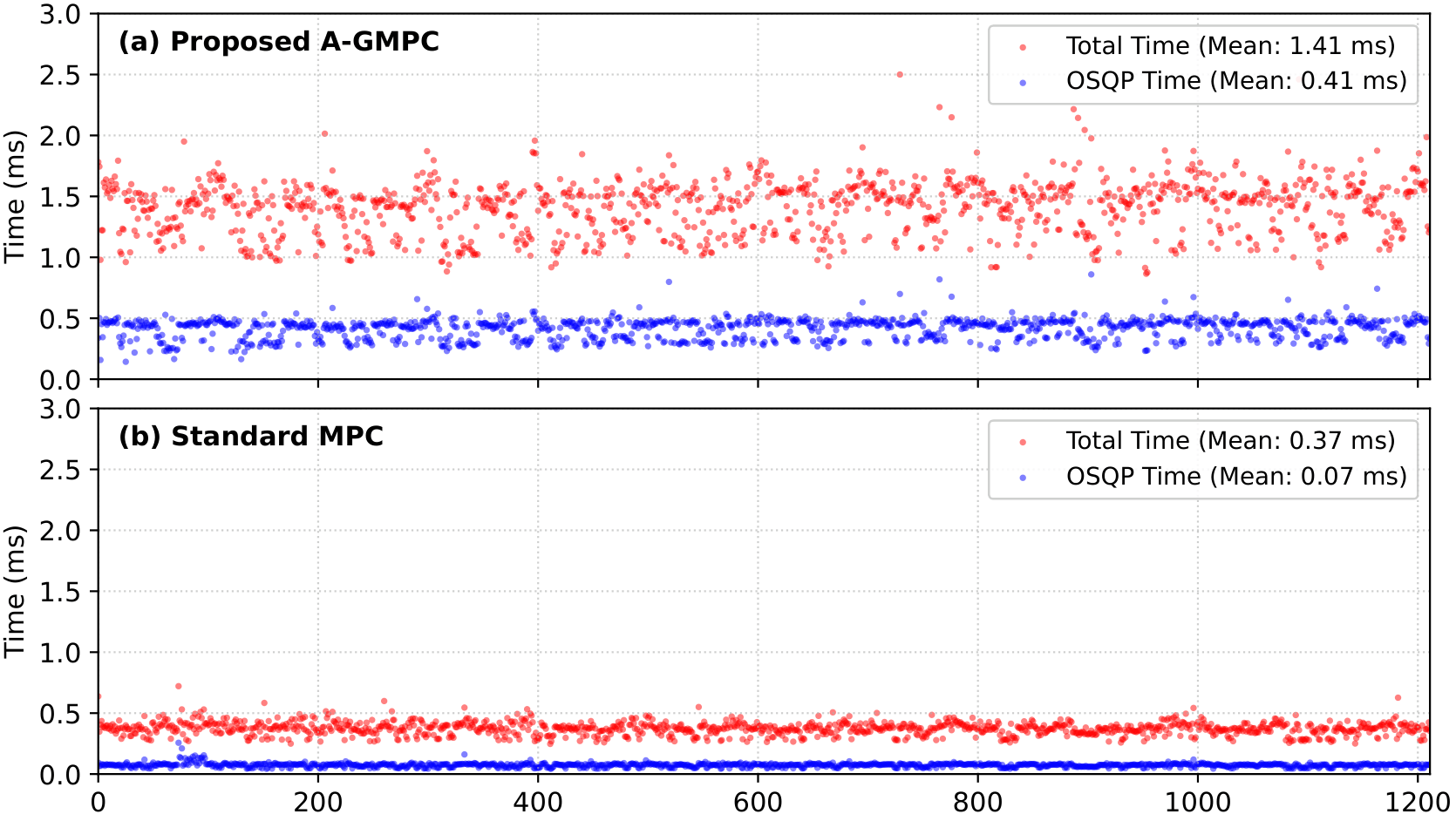}
        \vspace{-5mm}
		\caption{Runtime statistics comparison. Both execute well within the 30 Hz (33.3 ms) control loop.}
		\label{fig:comp_time_boxplot}
        \vspace{-7mm}
	\end{figure}
	
    \subsubsection{Real-World Path-Tracking Performance}
    
    The proposed architecture was deployed on the physical robot to verify sim-to-real consistency. Table~\ref{tab:real_tracking_rmse} summarizes the tracking accuracy at different average speeds. Fig.~\ref{fig:all_tracking_paths}(c)--(d) shows representative real-world tracking trajectories at approximately 1.1~m/s, while Fig.~\ref{fig:comp_time_boxplot} presents the corresponding runtime statistics.
    
    The results in Table~\ref{tab:real_tracking_rmse} show that A-GMPC consistently achieves lower RMSE than the standard MPC for all tested trajectories and speeds. The improvement is especially clear on the square trajectory, where sharp turns increase the tracking difficulty. These results confirm that the proposed controller improves real-world tracking accuracy while maintaining real-time performance.
    
    A supplementary video of the real-world path-tracking results is available at \url{https://youtu.be/eXZOD7MUVX8}.

    \begin{table}
        \centering
        \caption{RMSE Comparison in Real-World Deployment at Different Average Speeds (Unit [m]). The Best Results are in Bold.}
        \label{tab:real_tracking_rmse}
        \footnotesize
        \setlength{\tabcolsep}{4pt}
        \renewcommand{\arraystretch}{1.2}
    
        \begin{tabularx}{0.92\linewidth}{l l >{\centering\arraybackslash}X >{\centering\arraybackslash}X}
            \toprule
            \makecell{\textbf{Average} \\ \textbf{Speed}} &
            \textbf{Trajectory} &
            \makecell{\textbf{Standard}\\\textbf{MPC}} &
            \makecell{\textbf{Proposed}\\\textbf{A-GMPC}} \\
            \midrule
    
            \multirow{2}{*}{$0.6$ m/s}
            & Lemniscate & 0.3194 & \textbf{0.2349} \\
            & Square     & 0.4260 & \textbf{0.2588} \\
            \midrule
    
            \multirow{2}{*}{$1.1$ m/s}
            & Lemniscate & 0.3281 & \textbf{0.3048} \\
            & Square     & 0.4413 & \textbf{0.2777} \\
            \midrule
    
            \multirow{2}{*}{$1.5$ m/s}
            & Lemniscate & 0.5149 & \textbf{0.3582} \\
            & Square     & 1.1475 & \textbf{0.6182} \\
            \bottomrule
        \end{tabularx}
        \vspace{-7mm}
    \end{table}
	
	\subsubsection{Autonomous Navigation Demonstration}

\begin{figure}[h]
		\centering
		\includegraphics[width=0.48\linewidth,height=3.2cm]{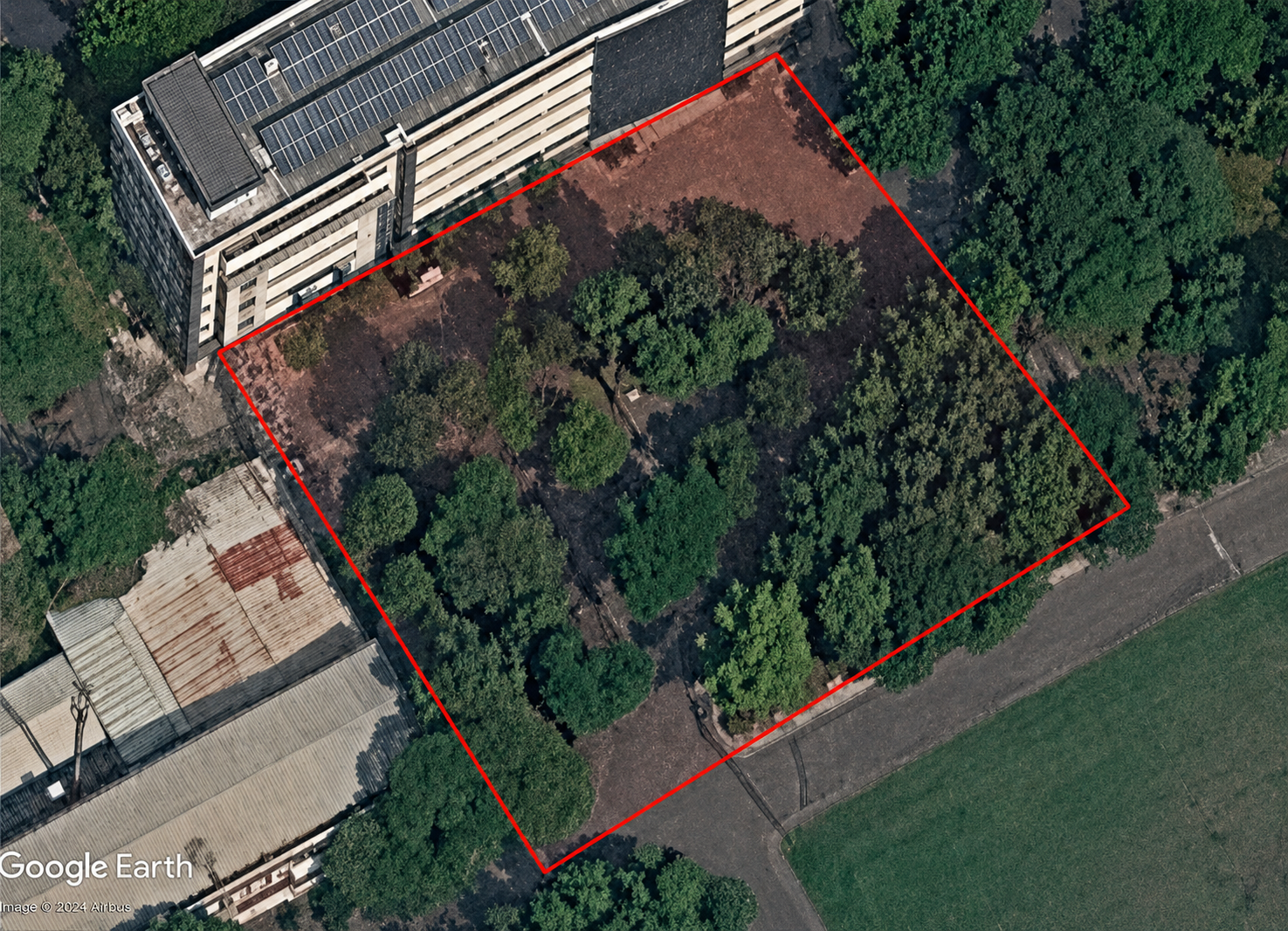}
		\includegraphics[width=0.48\linewidth,height=3.2cm]{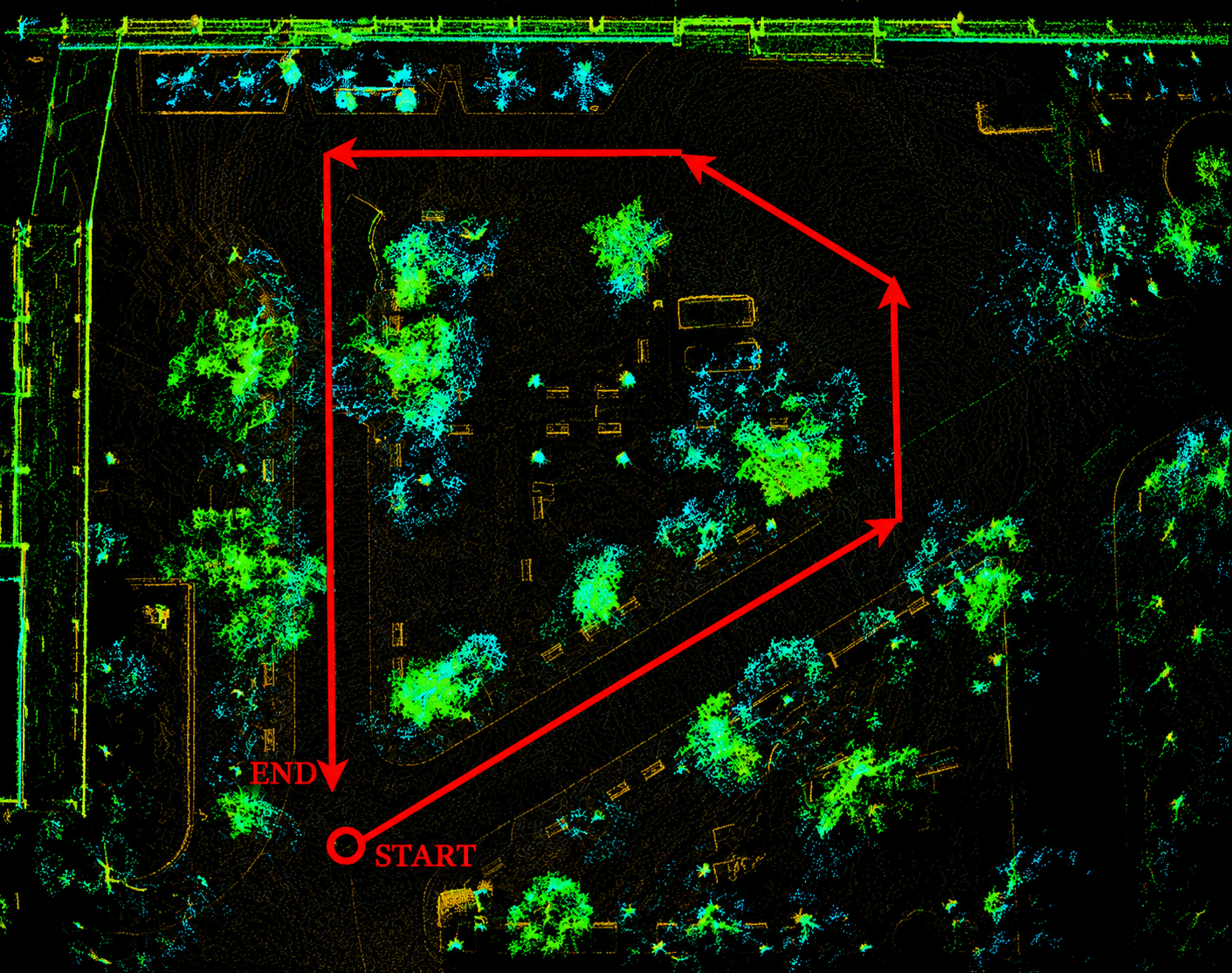}
		\caption{End-to-end real-world navigation demonstration with environmental mapping. Left: satellite view of the experimental site. Right: registered 3D point-cloud map reconstructed by the onboard perception module during operation. The reconstructed map is subsequently converted into a 2D occupancy representation for global path planning.}
		\label{fig:end_to_end_map}
        \vspace{-2mm}
	\end{figure}

\begin{figure}[!t] 
        \centering
        \vspace{-10mm}
        \includegraphics[width=\linewidth]{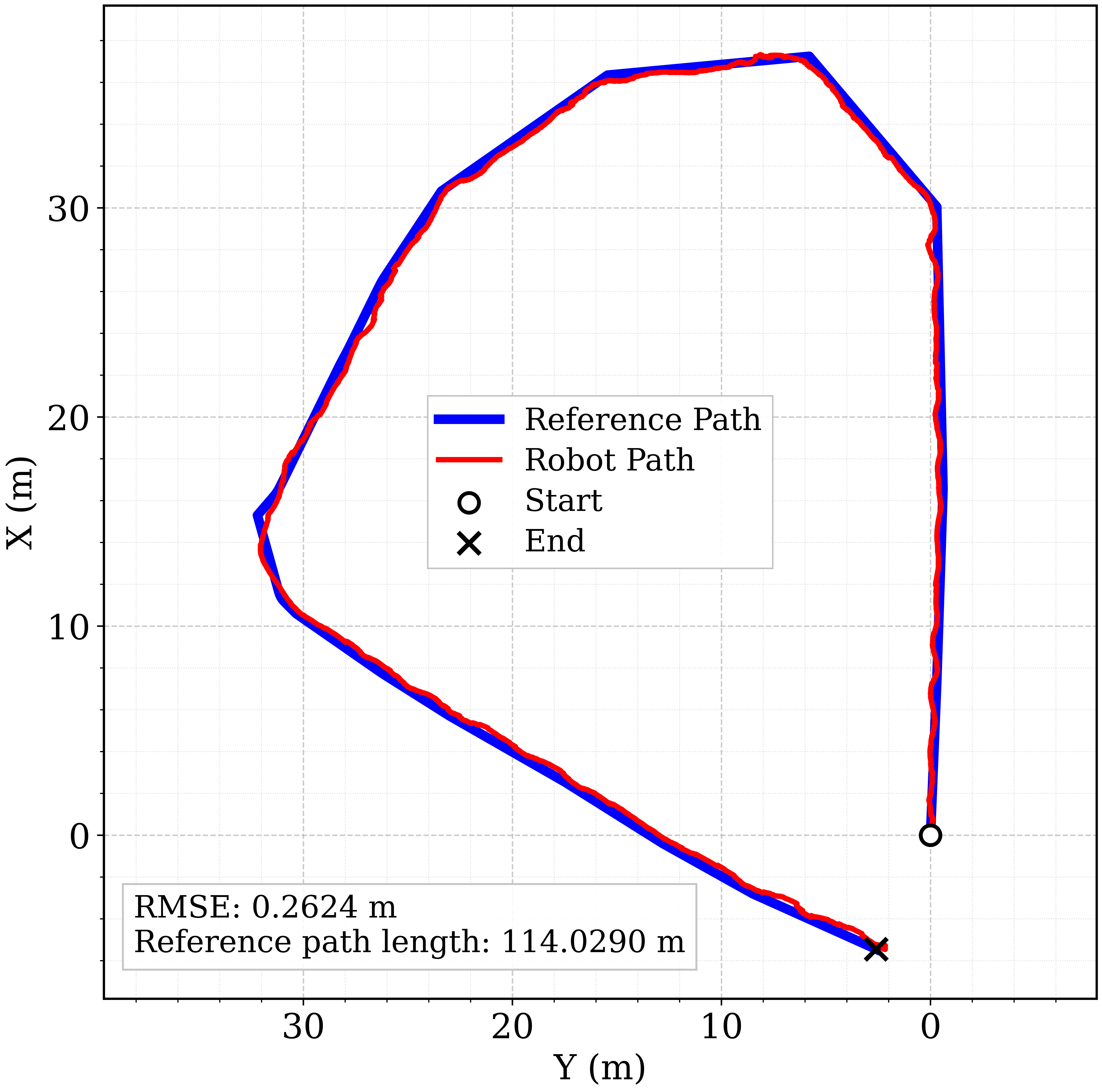}
        \caption{Trajectory tracking performance during the end-to-end real-world navigation experiment, illustrating the reference path and the actual executed robot path.}
        \label{fig:end2end_tracking}
        \vspace{-5mm}
    \end{figure}

	An end-to-end real-world navigation experiment was conducted to validate the complete autonomy pipeline. Figure~\ref{fig:end_to_end_map} shows the experimental site and the corresponding 3D point-cloud map reconstructed during operation, which was later converted into a 2D occupancy map for global planning. 
	The results confirm that the proposed system can integrate onboard perception, mapping, path planning, and trajectory tracking in a unified real-world framework (see Fig.~\ref{fig:end2end_tracking}). A supplementary video is available at \url{https://youtu.be/2TXuBDscRR4}.
    
	\section{Conclusion}
	
	This paper presented a simulation-to-real autonomous navigation system for an Ackermann-steered mobile robot built on an existing mechanical platform. 
	The proposed framework integrates onboard localization, path planning, and trajectory tracking into a unified pipeline for autonomous navigation. 
	Simulation and real-world experiments demonstrated that the system can generate smooth and safe paths while achieving reliable tracking performance, confirming the feasibility of sim-to-real deployment. 
	Future work will focus on more robust tracking of humans and moving objects, as well as navigation in more complex and dynamic outdoor environments.
		
	\section*{Acknowledgment}
	The authors acknowledge the support and facilities provided by Ho Chi Minh City University of Technology (HCMUT), VNU-HCM, for this study.
		
		\bibliographystyle{IEEEtran}
		\bibliography{ref}

@INPROCEEDINGS{wiedemann2024simmodel,
	author={Wiedemann, Marvin and Ahmed, Ossama and Dieckh{\"o}fer, Anna and Gasoto, Renato and Kerner, S{\"o}ren},
	booktitle={2024 IEEE International Conference on Robotics and Automation (ICRA)},
	title={Simulation Modeling of Highly Dynamic Omnidirectional Mobile Robots Based on Real-World Data},
	year={2024},
	pages={16923-16929}
}

@ARTICLE{tang2021geometric,
	author={Tang, Gang and Tang, Congqiang and Claramunt, Christophe and Hu, Xiong and Zhou, Peipei},
	journal={IEEE Access}, 
	title={Geometric A-Star Algorithm: An Improved A-Star Algorithm for AGV Path Planning in a Port Environment}, 
	year={2021},
	volume={9},
	number={},
	pages={59196-59210}
	}

@InProceedings{moore2016ekf,
	author="Moore, Thomas
	and Stouch, Daniel",
	editor="Menegatti, Emanuele
	and Michael, Nathan
	and Berns, Karsten
	and Yamaguchi, Hiroaki",
	title="A Generalized Extended Kalman Filter Implementation for the Robot Operating System",
	booktitle="Intelligent Autonomous Systems 13",
	year="2016",
	publisher="Springer International Publishing",
	address="Cham",
	pages="335--348",
}

@ARTICLE{hart1968astar,
	author={Hart, Peter E. and Nilsson, Nils J. and Raphael, Bertram},
	journal={IEEE Transactions on Systems Science and Cybernetics}, 
	title={A Formal Basis for the Heuristic Determination of Minimum Cost Paths}, 
	year={1968},
	volume={4},
	number={2},
	pages={100-107}
	}

@ARTICLE{kim2022opensource_nav,
	author={Kim, Taekyung and Lim, Seunghyun and Shin, Gwanjun and Sim, Geonhee and Yun, Dongwon},
	journal={IEEE Access},
	title={An Open-Source Low-Cost Mobile Robot System with an RGB-D Camera and Efficient Real-Time Navigation Algorithm},
	year={2022},
	volume={10},
	pages={127871-127881}
}

@ARTICLE{yu2018bus_nav,
	author={Yu, Lingli and Kong, Decheng and Shao, Xuanya and Yan, Xiaoxin},
	journal={IEEE Access},
	title={A Path Planning and Navigation Control System Design for Driverless Electric Bus},
	year={2018},
	volume={6},
	pages={53960-53975}
}

@ARTICLE{lin2024astar,
	author={Lin, Zhi and Wu, Kang and Shen, Rulin and Yu, Xin and Huang, Shiquan},
	journal={IEEE Transactions on Vehicular Technology}, 
	title={An Efficient and Accurate A-Star Algorithm for Autonomous Vehicle Path Planning}, 
	year={2024},
	volume={73},
	number={6},
	pages={9003-9008}
}

@ARTICLE{tang2024gmpc,
	author={Tang, Jiawei and Wu, Shuang and Lan, Bo and Dong, Yahui and Jin, Yuqiang and Tian, Guangjian and Zhang, Wen-An and Shi, Ling},
	journal={IEEE Robotics and Automation Letters}, 
	title={GMPC: Geometric Model Predictive Control for Wheeled Mobile Robot Trajectory Tracking}, 
	year={2024},
	volume={9},
	number={5},
	pages={4822-4829}
}

@INPROCEEDINGS{le2024mpc,
	author={Le, Cong-Tu and Nguyen, Nhat-Hung and Nguyen, Vinh-Hao},
	booktitle={2024 International Conference on Advanced Technologies for Communications (ATC)}, 
	title={Model Predictive Control based Trajectory Tracking for Autonomous Vehicles}, 
	year={2024},
	volume={},
	number={},
	pages={124-129}
}

@INPROCEEDINGS{jian2023mpc,
	author={Jian, Zhuozhu and Yan, Zihong and Lei, Xuanang and Lu, Zihong and Lan, Bin and Wang, Xueqian and Liang, Bin},
	booktitle={2023 IEEE International Conference on Robotics and Automation (ICRA)},
	title={Dynamic Control Barrier Function-based Model Predictive Control to Safety-Critical Obstacle-Avoidance of Mobile Robot},
	year={2023},
	pages={3679-3685}
}

@ARTICLE{lu2023ommpc,
	author={Lu, Guozheng and Xu, Wei and Zhang, Fu},
	journal={IEEE Transactions on Industrial Electronics}, 
	title={On-Manifold Model Predictive Control for Trajectory Tracking on Robotic Systems}, 
	year={2023},
	volume={70},
	number={9},
	pages={9192-9202}
}

@ARTICLE{song2023isolating,
	author={Song, Jiarui and Tao, Gang and Zang, Zheng and Dong, Haotian and Wang, Boyang and Gong, Jianwei},
	journal={IEEE Robotics and Automation Letters},
	title={Isolating Trajectory Tracking From Motion Control: A Model Predictive Control and Robust Control Framework for Unmanned Ground Vehicles},
	year={2023},
	volume={8},
	number={3},
	pages={1699--1706},
	doi={10.1109/LRA.2023.3242151}
}

@ARTICLE{zheng2025fastlivo2,
	author={Zheng, Chunran and Xu, Wei and Zou, Zuhao and Hua, Tong and Yuan, Chongjian and He, Dongjiao and Zhou, Bingyang and Liu, Zheng and Lin, Jiarong and Zhu, Fangcheng and Ren, Yunfan and Wang, Rong and Meng, Fanle and Zhang, Fu},
	journal={IEEE Transactions on Robotics}, 
	title={FAST-LIVO2: Fast, Direct LiDAR–Inertial–Visual Odometry}, 
	year={2025},
	volume={41},
	number={},
	pages={326-346}
}

@ARTICLE{xu2022fastlio2,
	author={Xu, Wei and Cai, Yixi and He, Dongjiao and Lin, Jiarong and Zhang, Fu},
	journal={IEEE Transactions on Robotics}, 
	title={FAST-LIO2: Fast Direct LiDAR-Inertial Odometry}, 
	year={2022},
	volume={38},
	number={4},
	pages={2053-2073}
}

@ARTICLE{nguyen2023slict,
	author={Nguyen, Thien-Minh and Duberg, Daniel and Jensfelt, Patric and Yuan, Shenghai and Xie, Lihua},
	journal={IEEE Robotics and Automation Letters}, 
	title={SLICT: Multi-Input Multi-Scale Surfel-Based Lidar-Inertial Continuous-Time Odometry and Mapping}, 
	year={2023},
	volume={8},
	number={4},
	pages={2102-2109}
}
		
	\end{document}